\documentclass[final,5p,times,twocolumn]{elsarticle}

\usepackage{amssymb}
\usepackage{amsmath}
\usepackage{siunitx}
\usepackage{multirow}
\usepackage{tabularx,booktabs}
\usepackage[table,xcdraw]{xcolor}

\usepackage{lineno}

\journal{International Journal of Computer Vision}

\begin{document}

\begin{frontmatter}



\title{Neuro‑Inspired Visual Pattern Recognition via Biological Reservoir Computing}


\author[label1]{Luca Ciampi\corref{cor1}} 
\ead{luca.ciampi@isti.cnr.it}
\author[label1]{Ludovico Iannello\corref{cor1}}
\ead{ludovico.iannello@isti.cnr.it}
\author[label2]{Fabrizio Tonelli\corref{cor1}}
\ead{fabrizio.tonelli@sns.it}
\author[label1]{Gabriele Lagani}
\author[label3]{Angelo Di Garbo}
\author[label2]{Federico Cremisi}
\author[label1]{Giuseppe Amato}

\cortext[cor1]{Corresponding authors. They contribute equally to this work.}
\affiliation[label1]{organization={Institute of Information Science and Technologies of the National Research Council of Italy (ISTI-CNR)},
            city={Pisa},
            country={Italy}}
\affiliation[label2]{organization={Biology Laboratory at Scuola Normale Superiore (Bio@SNS)},
            city={Pisa},
            country={Italy}}
\affiliation[label3]{organization={Institute of Biophysics of the National Research Council of Italy (IBF-CNR)},
            city={Pisa},
            country={Italy}}

\begin{abstract}
In this paper, we present a neuro-inspired approach to reservoir computing (RC) in which a network of \textit{in vitro} cultured cortical neurons serves as the physical reservoir. Rather than relying on artificial recurrent models to approximate neural dynamics, our biological reservoir computing (BRC) system leverages the spontaneous and stimulus‑evoked activity of living neural circuits as its computational substrate. A high-density multi-electrode array (HD-MEA) provides simultaneous stimulation and readout across hundreds of channels: input patterns are delivered through selected electrodes, while the remaining ones capture the resulting high-dimensional neural responses, yielding a biologically grounded feature representation. A linear readout layer (single-layer perceptron) is then trained to classify these reservoir states, enabling the living neural network to perform static visual pattern-recognition tasks within a computer-vision framework.
We evaluate the system across a sequence of tasks of increasing difficulty, ranging from pointwise stimuli to oriented bars, clock‑digit‑like shapes, and handwritten digits from the MNIST dataset. Despite the inherent variability of biological neural responses---arising from noise, spontaneous activity, and inter‑session differences---the system consistently generates high‑dimensional representations that support accurate classification. These results demonstrate that \textit{in vitro} cortical networks can function as effective reservoirs for static visual pattern recognition, opening new avenues for integrating living neural substrates into neuromorphic computing frameworks. More broadly, this work contributes to the effort to incorporate biological principles into machine learning and supports the goals of neuro‑inspired vision by illustrating how living neural systems can inform the design of efficient and biologically grounded computational models.

\end{abstract}




\end{frontmatter}


\section{Introduction}

\label{sec:intro}
Reservoir computing (RC)~\cite{DBLP:journals/csr/LukoseviciusJ09} is a neural computing paradigm originally developed to process temporal signals, such as time-series~\cite{DBLP:journals/tnn/BianchiSLJ21,DBLP:conf/ijcnn/KatoTNH22} and speech~\cite{DBLP:journals/ficn/YonemuraK24,DBLP:journals/tnn/ZhangLJC15}. It operates by feeding input data into a high-dimensional recurrent system---known as the reservoir---whose dynamic response projects the input into a nonlinear latent space. This transformation often yields representations that are more easily separable by simple classifiers, enabling efficient learning with minimal training. In fact, only the readout layer---which maps the reservoir states to the final output---is trained in a supervised way, while the reservoir itself exploits intrinsic recurrent dynamics, foregoing weight optimization within the recurrent layer. These dynamics---either fixed or shaped by unsupervised processes---are responsible for mapping inputs into expressive latent representations.
Although RC was originally designed for sequential data~\cite{DBLP:journals/tnn/BianchiSLJ21,DBLP:conf/ijcnn/KatoTNH22,DBLP:journals/ficn/YonemuraK24,DBLP:journals/tnn/ZhangLJC15}, recent studies have demonstrated its effectiveness in static pattern recognition tasks, such as image classification, by leveraging the transient dynamics elicited by spatially structured stimuli~\cite{DBLP:conf/icpr/WangL16,9339389}, even though this application scenario remains relatively underexplored.

\begin{figure*}[t]
    \centerline{\includegraphics[width=1\linewidth]{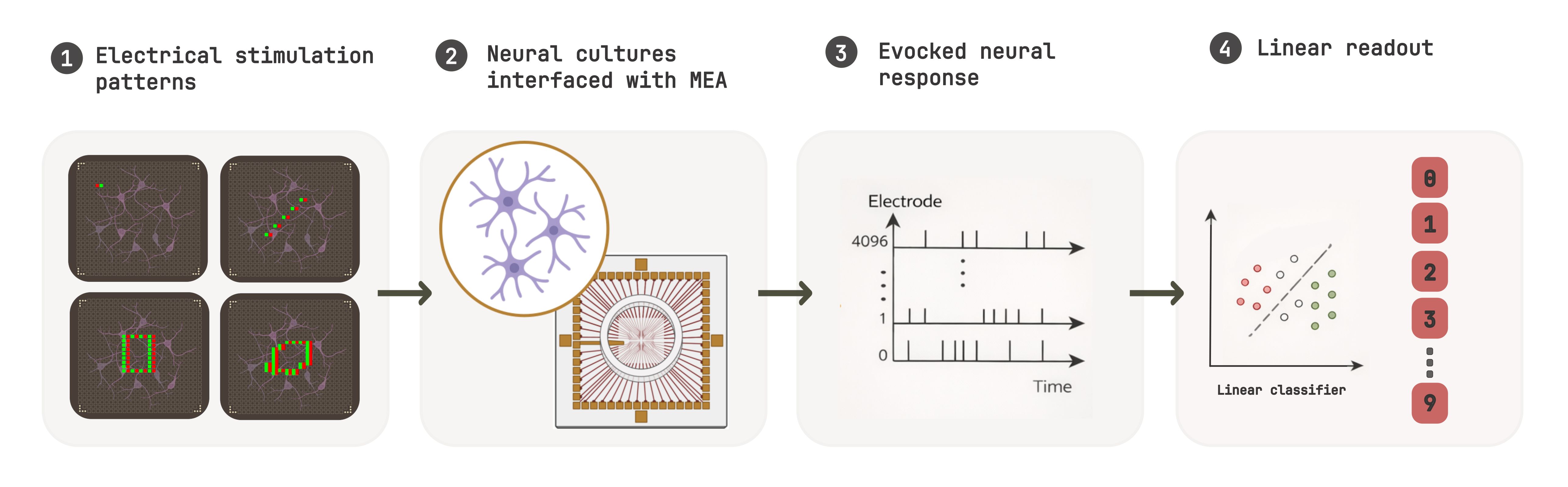}}
    \caption{
    \textbf{Schematic overview of our biological reservoir computing (BRC) framework.} We introduce a system that encodes spatial stimulation patterns into high‑dimensional neural representations generated by a living cortical network, enabling biologically grounded visual pattern recognition.
    (1) Input symbols are encoded as spatially distributed electrical stimulation patterns delivered through selected electrodes of a high‑density multi‑electrode array (HD‑MEA). 
    (2) These patterns are delivered to a cultured cortical network that has been plated onto the HD‑MEA surface, allowing precise electrical stimulation and high‑resolution recording of the resulting activity. 
    (3) The evoked neural responses are captured on dedicated recording channels and encoded into high‑dimensional reservoir states that reflect the nonlinear and transient dynamics of the biological substrate. 
    (4) A linear readout module (single‑layer perceptron) is trained to map these reservoir states to their corresponding input classes, allowing the living neural network to function as a physical reservoir for static pattern‑recognition tasks.
    }
    \label{fig:MEA_overview}
\end{figure*}

Among the most widely adopted RC models are the echo state network (ESN)~\cite{DBLP:journals/ijon/ValenciaVF23,DBLP:journals/nn/GallicchioMP18,DBLP:journals/corr/abs-1712-04323,DBLP:conf/ijcnn/KatoTNH22} and the liquid state machine (LSM)~\cite{DBLP:journals/neco/MaassNM02,DBLP:journals/tnn/ZhangLJC15,9339389}. ESNs consist of randomly connected continuous-valued recurrent units, while LSMs employ spiking neurons to emulate biologically inspired temporal dynamics. Despite their architectural differences, both models share the core principle of relying on fixed intrinsic recurrent dynamics to enrich input representations, followed by a trained readout layer for task-specific learning. However, despite their success, these models usually rely on artificial implementations of neural activity and remain abstract approximations of biological neural processes~\cite{DBLP:journals/ijon/ValenciaVF23,DBLP:journals/nn/GallicchioMP18,DBLP:journals/corr/abs-1712-04323,DBLP:conf/ijcnn/KatoTNH22,DBLP:journals/neco/MaassNM02,DBLP:journals/tnn/ZhangLJC15,9339389}.

In this work, we present a \textit{biologically} grounded instantiation of the RC paradigm, where the reservoir is not simulated but physically realized through a living network of cultured cortical neurons---a framework we refer to as \textit{biological reservoir computing} (BRC). Unlike conventional RC systems based on artificial neuron models, our approach leverages the intrinsic dynamics of real neuronal activity to project input stimuli into a high-dimensional feature space. By empirically evaluating the feasibility and performance of BRC, we investigate its applicability to static pattern recognition tasks, where spatially encoded stimuli elicit transient neural responses used for classification. This contributes to the broader effort of integrating biological principles into machine learning, in line with neuroscientific evidence on the role of transient dynamics in information processing~\cite{https://doi.org/10.1111/j.1460-9568.2007.05976.x,DRISCOLL2018873}. Furthermore, by offloading part of the computation to a physical neural substrate, BRC may offer advantages in energy efficiency---an increasingly critical concern in scaling modern AI~\cite{DBLP:conf/aiia/BadarVSGMIAZ21,vanWynsberghe2021SustainableAA,DBLP:journals/ijon/BolonCanedoMCA24}. 

Specifically, our BRC system is implemented using a high-density multi-electrode array (HD-MEA)~\cite{bonifazi}, which enables both electrical stimulation and high-resolution recording of neural activity. Cortical neurons are derived \textit{in vitro} from mouse embryonic stem cells (mESCs) using established differentiation protocols~\cite{fabri, terrigno_neurons_2018}, and once matured, they form spontaneously active networks~\cite{fabri, iannello_criticality_2025}, which serve as the biological substrate of our reservoir. Input patterns are encoded as spatially distributed stimulation sequences, mimicking visual stimuli, and delivered to the culture via the HD-MEA. The resulting neural responses are recorded and transformed into feature vectors for downstream pattern classification, which is performed by a simple readout layer consisting of a single-layer perceptron. A visual overview of our approach is shown in Fig.~\ref{fig:MEA_overview}.

To evaluate the ability of our BRC system to generate discriminative feature representations across increasing levels of input complexity, we designed a progressive experimental study spanning tasks of growing difficulty. We began with elementary pointwise stimuli, serving as baseline inputs to probe the responsiveness of the system. We then introduced structured geometric patterns, such as bars with varying orientations, to assess the capacity of the reservoir to encode spatial relationships. Next, we tested the classification of ten clock-digit-like spatial input patterns (from 0 to 9).
Finally, we challenged the system with a more demanding task: the classification of handwritten digits from the MNIST dataset~\cite{726791}, where each image was mapped to a spatial stimulation sequence, allowing us to evaluate the performance of the system on real-world visual inputs. Despite the inherent variability in biological responses---stemming from noise, spontaneous activity, and inter-session differences---the system consistently produced high-dimensional representations that enabled classification. These findings demonstrate that cortical neuronal networks cultured \textit{in vitro} can serve as effective reservoirs for static pattern recognition, opening new avenues for integrating living neural substrates into neuromorphic computing frameworks. In this sense, our study provides a concrete biological instantiation of neuro-inspired vision, demonstrating how living neural circuits can perform core perceptual computations traditionally implemented in artificial models.

Concretely, this paper offers the following key contributions:
\begin{itemize} 
    \item We introduce a biologically-grounded reservoir computing paradigm (BRC), in which high-dimensional feature representations are generated from the intrinsic dynamics of cortical neuronal networks cultured \textit{in vitro}. The system interfaces with the biological substrate via a high-density multi-electrode array (HD-MEA), enabling precise electrical stimulation and high-resolution recording of neural responses. 
    \item We define and calibrate a stimulation protocol tailored for HD-MEA-based interaction with neuronal cultures, ensuring consistent and robust evoked responses across varying spatial stimulation patterns and experimental sessions.
    \item We conduct a comprehensive experimental evaluation of the BRC system across a series of static pattern recognition tasks with increasing complexity. Starting from simple pointwise stimuli, we progress to geometric patterns (e.g., oriented bars), then to clock-digit-like spatial configurations (0–9), and finally to real-world handwritten digits from the MNIST dataset. This progression allows us to systematically assess the discriminative power of biologically derived features under diverse input conditions. 
\end{itemize}

This work substantially extends our previous preliminary works~\cite{Iannello_2025,10350607}, which provided initial evidence supporting our approach, albeit using a limited set of spatial patterns and a restricted scope of analysis. In contrast, the present work introduces several major advancements: (i) significantly broadening the experimental scope by incorporating a comprehensive evaluation on the MNIST dataset, thus extending the range of experiments beyond the restricted patterns previously considered and enabling assessment on real-world handwritten visual inputs; (ii) an in-depth and systematic analysis of the obtained results, uncovering new insights into system behavior; and (iii) a thorough characterization of the biological cultures, integrating perspectives from both neuroscience and biology.

The remainder of this manuscript is organized as follows. Section~\ref{sec:rel_work} reviews related work in reservoir computing and neuro-inspired computing. Section~\ref{sec:method} describes the proposed methodology, including neurobiological protocols and HD-MEA interfacing. Section~\ref{sec:exp} presents the experimental setup, along with results and analysis. Finally, Sec.~\ref{sec:conclusions} summarizes our findings and outlines future research directions.

\section{Related Works}
\label{sec:rel_work}
A longstanding question in deep learning research concerns the degree to which current computational models align with biological neural systems. Numerous contributions have critically evaluated how artificial architectures reflect the structural and functional complexity of the brain~\cite{DBLP:journals/ficn/MarblestoneWK16,Hassabis_2017,f3a024ff8e474a1ea4e798aaa5536830,DBLP:conf/atal/Tenenbaum18,Zador2019ACO}. This has led to a growing interest in biologically inspired alternatives, aimed at enhancing both machine learning capabilities and cognitive modeling~\cite{DBLP:journals/corr/abs-2307-16236,DBLP:journals/cogsr/Luczak25,DBLP:journals/ojcands/ParhiU20,DBLP:journals/corr/abs-2412-03192,DBLP:conf/eccv/CiampiLAF24,DBLP:journals/ijon/LaganiFGFA24,DBLP:journals/nn/LaganiFGA21}.
Such efforts have produced a wide variety of models that attempt to narrow the gap between artificial and biological intelligence. Some approaches focus on implementing neural computations that are more biologically grounded~\cite{DBLP:journals/ficn/Diehl015,DBLP:journals/ficn/FerreMT18,DBLP:journals/cogcom/SunCCS23}, while others refine the mechanisms of synaptic transmission and learning~\cite{DBLP:journals/nn/IllingGB19,DBLP:journals/nca/LaganiFGA22,DBLP:conf/iclr/JourneRGM23}.

In this broader effort to bridge artificial and biological intelligence, reservoir computing (RC)~\cite{DBLP:journals/csr/LukoseviciusJ09} has attracted attention as a framework for capturing complex neural dynamics through recurrent architectures. Two notable RC variants are echo state networks (ESNs)~\cite{DBLP:journals/ijon/ValenciaVF23,DBLP:journals/nn/GallicchioMP18,DBLP:journals/corr/abs-1712-04323,DBLP:conf/ijcnn/KatoTNH22} and liquid state machines (LSMs)~\cite{DBLP:journals/neco/MaassNM02,DBLP:journals/tnn/ZhangLJC15,9339389}, which differ in their computational principles and in the extent to which they reflect biological processes. LSMs typically employ spiking neuron models~\cite{DBLP:books/cu/GerstnerK02} to generate rich temporal dynamics, while ESNs rely on large recurrent networks of randomly connected continuous-valued units to embed input signals into high-dimensional spaces suitable for classification.
Although ESNs are initialized stochastically, specific design strategies---such as spectral radius tuning and input scaling---are commonly adopted to ensure that the network operates within a stable and computationally effective regime~\cite{DBLP:journals/nn/GallicchioMP18,DBLP:series/lncs/Lukosevicius12,DBLP:journals/tnn/SteinerJB23}. Similar considerations apply to LSMs, where appropriate parameter choices are crucial to achieving meaningful and robust dynamics~\cite{DBLP:conf/ijcnn/GrzybCWK09a,DBLP:journals/neco/MaassNM02}.
To further enhance biological plausibility, several extensions of RC have been proposed. Spiking variants that incorporate spike-timing-dependent plasticity (STDP) and intrinsic plasticity aim to emulate biologically grounded learning mechanisms~\cite{Song_2000,DBLP:books/cu/GerstnerK02}, whereas gating mechanisms have been introduced in ESNs to improve the handling of long-term dependencies in nonlinear recurrent systems~\cite{DBLP:conf/inista/SarliGM20}. These biologically inspired RC-based approaches have demonstrated promising results in sequential data domains such as speech recognition~\cite{DBLP:journals/ficn/YonemuraK24,DBLP:journals/tnn/ZhangLJC15}, time-series forecasting~\cite{DBLP:journals/tnn/BianchiSLJ21,DBLP:conf/ijcnn/KatoTNH22}, and continual learning~\cite{DBLP:conf/esann/CossuBCGL21}. To a lesser extent, they have also been applied to static pattern recognition tasks~\cite{DBLP:conf/icpr/WangL16,9339389}.

In contrast to most prior approaches, this study enhances the biological plausibility of RC by leveraging a living network of cultured cortical neurons as the computational reservoir. Specifically, the concept we refer to as biological reservoir computing (BRC) involves interfacing cultured neuronal populations with a high-density multi-electrode array (HD-MEA) device, enabling the biological network to serve as the dynamic reservoir substrate.
While several prior studies have explored HD-MEA-based interfacing with biological neurons~\cite{Shahaf8782,DBLP:journals/tbe/RuaroBT05,DBLP:journals/ijon/FerrandezLPF13,GOEL2016320,DBLP:journals/ploscb/PastoreMGM18}, only a few have investigated their use within RC frameworks. A few previous efforts have also employed biological neurons in RC settings; however, they either focused on input separation via distinct stimuli~\cite{DOCKENDORF200990}, or---as in the work by Cai et al.~\cite{cai2023brain}---used BRC to demonstrate that brain organoids can serve as biological reservoirs for speech recognition, leveraging the rich temporal dynamics of 3D neural structures to encode and process input sequences. Other works instead adopted alternative interfacing techniques such as optogenetics and calcium imaging~\cite{Sumi202306}.
In contrast, our approach centers on using BRC interfaced with HD-MEAs for static pattern recognition. Specifically, we apply spatially patterned electrical stimulation across HD-MEA channels (electrodes) without temporal sequencing, targeting static pattern recognition tasks. This strategy offers a promising direction for integrating biological computation with machine learning methodologies, potentially enabling improved energy efficiency and deeper insights into the dynamics of biological neural systems.


\section{Biological Reservoir Computing Framework}
\label{sec:method}
In this section, we introduce the biological reservoir computing (BRC) framework developed in this study. We first provide a high‑level overview of the architecture, followed by detailed descriptions of the preparation of the \textit{in vitro} cortical neuronal cultures, the input‑to‑electrode mapping and stimulation protocol, the acquisition and preprocessing of neural activity, and finally the training and evaluation of the readout classifier.


\subsection{Overview}
\begin{figure}[t]
    \centerline{\includegraphics[width=0.6\linewidth]{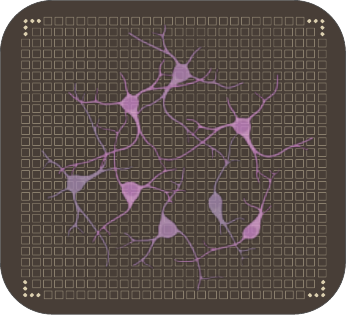}}
    \caption{\textbf{Schematic representation of a neuronal culture interfaced with an HD-MEA platform.} Each square corresponds to an individual electrode on the HD-MEA. Input patterns are encoded by mapping their elements to specific electrode positions. The remaining electrodes record the evoked spiking activity of the neuronal culture, which is then encoded into a high-dimensional feature representation within the biological reservoir. Neuron dimensions are enlarged for visualization purposes and are not to scale.}
    \label{fig:mea}
\end{figure}

Cultured neuronal networks grown \textit{in vitro} constitute complex dynamical systems of randomly interconnected processing units capable of generating rich, nonlinear activity. In our reservoir computing framework, these networks—interfaced with a high‑density multi‑electrode array (HD‑MEA), a bidirectional platform enabling both stimulation and electrophysiological recording (Fig.~\ref{fig:mea})—serve as the physical reservoir. A schematic of the overall pipeline is shown in Fig.~\ref{fig:MEA_overview}.
Input samples are mapped to specific electrode subsets and delivered as stimulation patterns. The evoked spiking activity is recorded from non-stimulated electrodes and aggregated into high-dimensional feature vectors representing the reservoir state. Under this configuration, the biological network functions as a feature extractor, transforming inputs into complex representations through its intrinsic dynamics. Finally, a single-layer perceptron acts as the downstream readout classifier to assess the discriminative power of these representations.


\subsection{In Vitro Cortical Neuronal Cultures 
}
To provide the biological substrate for our reservoir computing framework, cortical neuronal cultures were generated \textit{in vitro} following the MiBi protocol as previously described in~\cite{fabri}. Briefly, entorhinal-like neurons were derived from mouse embryonic stem cells (mESCs) using a temporally controlled differentiation process. The MiBi method (MAPK/ERK and BMP inhibition) involves transient and simultaneous inhibition of the MAPK/ERK and BMP signaling pathways in differentiating telencephalic progenitors during a narrow window of neural differentiation.

MiBi-derived neuronal cultures exhibit distinctive molecular and functional properties. At the molecular level, they display a gene expression profile closely matching that of the early postnatal entorhinal cortex~\cite{fabri}, a key region for spatial and contextual learning~\cite{malone_consistent_2024,buzsaki_memory_2013}. Functionally, these networks show high spontaneous activity characterized by synchronized spiking and bursting, as well as an elevated network burst rate (NBR). They operate in a state of self-organized criticality, evidenced by scale-invariant neuronal avalanches---a dynamic regime often associated with optimal computation~\cite{iannello_criticality_2025}. Among the cortical culture types studied by our group, MiBi networks exhibit the highest structural and functional connectivity~\cite{fabri}, making them particularly suitable for our reservoir computing paradigm.

At the end of embryonic neurogenesis---typically between days 20 and 25 of \textit{in vitro} differentiation (DIV20--25)---MiBi neuronal cultures were passaged and plated onto HD-MEA chips pre-coated with poly-ornithine and mouse laminin~\cite{fabri}. Cells were allowed to adhere and mature, with medium changes every two to three days. During maturation, cultures were continuously monitored to assess spontaneous activity and exclude anomalous samples or defective chips. Experimental stimulation was performed at least three weeks after plating (typically after DIV40).

\begin{figure*}[t]
    \centerline{\includegraphics[width=1\linewidth]{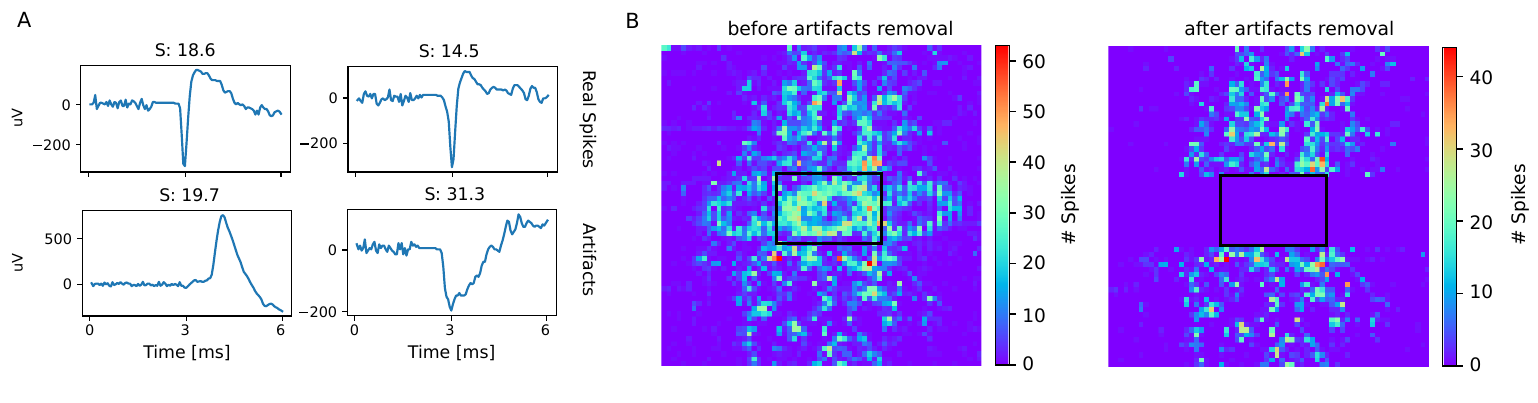}}
    \caption{
    \textbf{Artifact removal.} 
    In the second stage of postprocessing, stimulation‑induced electrical artifacts are identified and removed from the recorded signals. 
    (A) Examples of real neural spikes and stimulation‑evoked artifacts, with the normalized area $S$ reported above each trace; artifacts exhibit markedly larger $S$ values due to their broader temporal profiles. 
    (B) Heatmaps of activity accumulated over 20 repetitions of the same digit, shown before and after artifact removal. 
    Prior to filtering, the stimulated region (black rectangle) is dominated by artifacts; after removal, only biologically plausible spiking activity remains.
    }
    \label{fig:artifacts}
\end{figure*}

\subsection{Input Mapping and Stimulation Protocol}
In our reservoir computing framework, input samples are directly mapped onto specific subsets of HD-MEA electrodes. For instance, when an image is provided as input, its pixel grid of dimensions $H \times W$ is projected onto a corresponding $H \times W$ region of the electrode array. Each pixel is uniquely associated with an electrode, and its intensity is mapped to stimulation parameters---such as pulse amplitude and duration---applied at that electrode, which in turn stimulates the neurons cultured above the array.

Establishing a reliable interface with a biological neural network entails several technical complexities, particularly in calibrating stimulation parameters that must be carefully optimized to evoke consistent neural responses while preserving the integrity of the stimulation hardware. In this study, we present a systematic exploration and refinement of stimulation protocols tailored to activate cultured neuronal networks and enable stable electrophysiological recordings.

Stimulation is delivered via selected electrodes on the HD-MEA, configured in bipolar mode---each pair consisting of electrodes with opposite polarities. The stimuli are shaped as rectangular monophasic or biphasic waveforms, characterized by a defined amplitude (in $\si{\micro\ampere}$) and phase durations $\delta_+$ and $\delta_-$ (in $\si{\micro\second}$) corresponding to the positive and negative components, respectively. This configuration ensures a balanced current across the poles, contributing to long-term electrode stability. Given the distributed nature of current across multiple pairs, amplitude values are interpreted on a per-pair basis. The protocol design prioritizes a trade-off between stimulation efficacy and hardware preservation, avoiding excessive current or frequency that could lead to electrode degradation---an issue observed during preliminary high-intensity trials.

To minimize potential biases arising from temporal dependencies or sequence effects, stimulation patterns are randomized prior to delivery. A fixed inter-stimulus interval of $T = \SI{10}{\second}$ is enforced, allowing the network to return to baseline activity between stimuli and ensuring independence of evoked responses~\cite{DBLP:journals/tbe/RuaroBT05}.

The protocol was implemented on an HD-MEA provided by the 3Brain company\footnote{https://www.3brain.com/}using a custom Python script built on the manufacturer’s API. This script enables precise control over stimulation parameters, including randomized input sequencing, timing control, and electrode selection. The resulting framework ensures reproducibility and adaptability across experimental sessions. 


\subsection{Neural Response Acquisition and Preprocessing}
The acquisition pipeline consists of two main stages: spike detection and postprocessing. In the first stage, spike detection is performed in real time using a double-threshold algorithm designed for extracellular signals. The second stage involves artifact removal and the construction of high-dimensional activity vectors, as detailed below.

In the first stage, to characterize the response of the biological reservoir to external stimuli, we recorded neural activity from the HD-MEA within a symmetric time window spanning from $2 \si{\second}$ before to $2 \si{\second}$ after each stimulus onset. Spiking events were extracted in real time using a double-threshold detection algorithm relying on three key parameters: a sliding window of duration $w_s$, a low threshold $thr_l$, and a high threshold $thr_h$ for final spike confirmation. For each recording channel, candidate events are first identified as peaks exceeding $thr_l \times \sigma$, where $\sigma$ is the standard deviation of the full signal. To refine detection, segments crossing the low threshold are temporarily excluded, and a new standard deviation $\sigma_n$ is computed on the remaining signal. Spikes are then confirmed as local maxima (in absolute value) surpassing $thr_h \times \sigma_n$. 

In the second stage, postprocessing addresses artifacts introduced during stimulation. In fact, during stimulation, a significant number of electrical artifacts can be generated, which are distinct from real neural signals. To effectively isolate and remove these artifacts, a two-step filtering process is applied, leveraging the distinct characteristics of artifacts and spikes. First, we use a hard threshold based on amplitude. As artifacts often have abnormally high voltages (in absolute value), any signal exceeding a specific voltage threshold, such as $V_{thr} = 500 \mu V$ (empirically determined via comparative analysis), is automatically classified as an artifact and removed. Second, to account for artifacts that might not have a high amplitude but are unusually long in duration, we apply a second filter based on a normalized area defined as $S = \frac{\sum_t |V(t)|}{\max(|V(t)|)}$. This metric essentially measures the ``width'' of the signal relative to its height. Because artifacts are generally wider and longer-lasting than the short, transient neural spikes, they yield elevated normalized area values. By setting a second threshold $w_{thr} = 25$, we can filter out these longer and wider artifacts, ensuring that only electrical events with typical characteristics of neural spikes are preserved for analysis (see Fig.\ref{fig:artifacts}).

Finally, after artifact removal, the last step of postprocessing is to quantify the evoked activity. Let $t_s$ denote the stimulus onset time. For each electrode $(i, j)$, we quantify the evoked activity $A^W_{ij}(t_s)$ as the number of spikes detected within a post-stimulus time window $W$:

\begin{equation}
A^W_{ij}(t_s) = \sum_{t=t_s}^{t_s+W} s_{ij}(t),
\end{equation}

\noindent where $s_{ij}(t)$ is a binary indicator of spike occurrence at time $t$, and time is discretized according to the acquisition sampling rate of the system ($f_s = 20{,}000$ Hz). The resulting activity vector $A^W_{ij}(t_s)$ serves as the response of the biological reservoir. Each stimulus yields a 4096-dimensional feature vector, corresponding to the full electrode array.
To assess the ability of the network to propagate information beyond the stimulation site, we exclude a square region around the stimulated electrodes from the feature vector, thereby focusing on distal neural responses and the capacity of the reservoir for spatiotemporal integration.

\begin{figure*}[t]
    \centerline{\includegraphics[width=0.7 \linewidth]{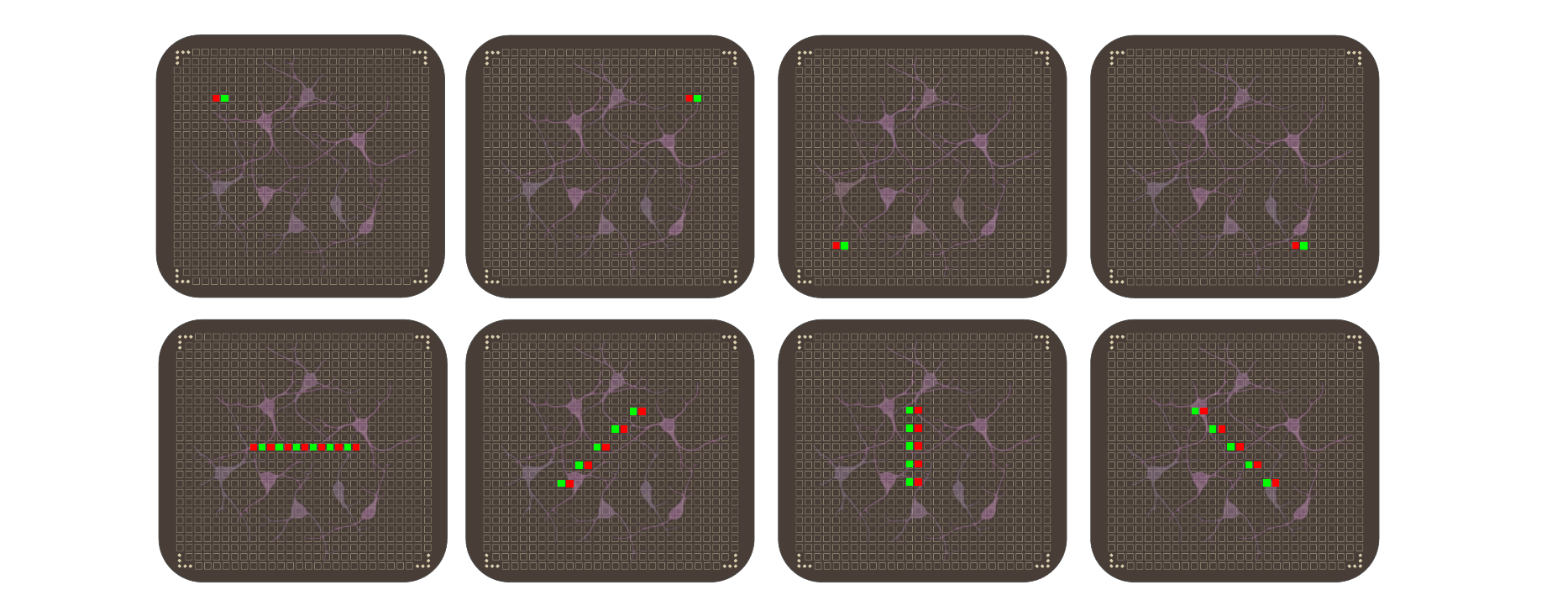}}
    \centerline{\includegraphics[width=0.7 \linewidth]{images_pdf/patterns_digit_pdf.pdf}}
    \centerline{\includegraphics[width=0.7 \linewidth]{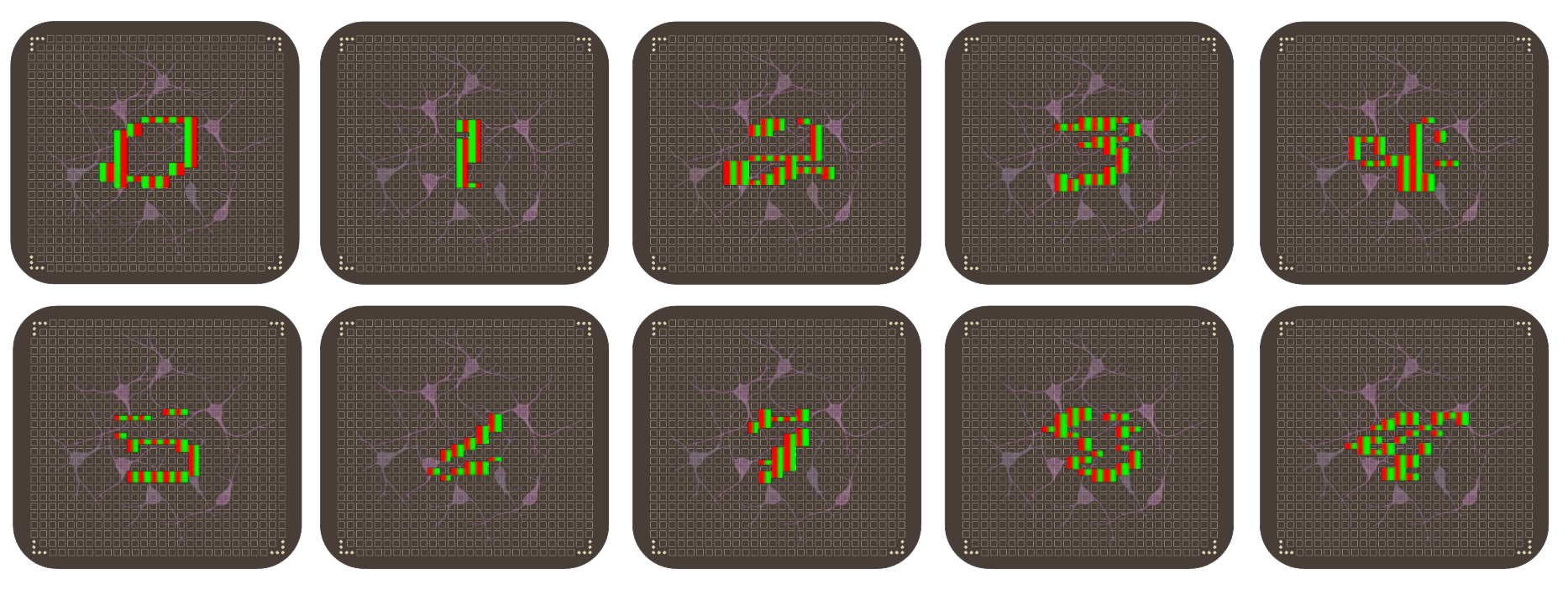}}
    \caption{\textbf{Visual representation of the input patterns across experiments.} Each panel shows the spatial layout of stimulation patterns mapped onto the MEA. The three sets of experiments include: (top) pointwise and oriented bars stimuli; (middle) clock-digit-like patterns; and (bottom) handwritten digits selected from subsets of MNIST patterns. Green and red pixels denote the positive and negative poles of the electrical stimulation, respectively.}
    \label{fig:input_patterns}
\end{figure*}

\subsection{Training of the Readout Classifier}
After preprocessing, the resulting feature vectors are used to train the single-layer perceptron (SLP) readout classifier. The training phase begins with defining a set of input patterns, each mapped to a specific subset of HD-MEA electrodes designated for stimulation. For each input, electrical pulses are delivered through the selected electrodes and repeated $N > 20$ times. This repetition is essential to account for the inherent variability of cultured neuronal networks, which exhibit spontaneous activity, intrinsic noise, and fluctuating dynamical states at the moment of stimulation. Consequently, identical input patterns can elicit diverse responses across trials.

After all responses are acquired, each trial is transformed into a high-dimensional feature vector representing spiking activity across the HD-MEA. These vectors are then labeled according to their corresponding input class. To train the classifier, the dataset is randomly partitioned into training and test sets, ensuring that samples from each class are evenly distributed.

Training is performed using a single-layer perceptron optimized via stochastic gradient descent (SGD)~\cite{DBLP:conf/compstat/Bottou10}, to minimize the cross-entropy loss~\cite{DBLP:journals/nca/KlineB05}. No validation set is employed, as no early stopping criterion is applied; this choice allows the full dataset to be utilized for both training and evaluation. The model is trained for 1,000 epochs with an adaptive learning rate, ensuring convergence and reproducibility across experimental sessions.

\subsection{Evaluation of the Readout Classifier}
After training, we evaluate the generalization capability of the readout classifier using a standard 5-fold cross-validation procedure. The dataset is randomly partitioned into five equally sized subsets. In each fold, four subsets are used to train the linear classifier, while the remaining one is reserved for testing. This process is repeated five times, ensuring that each subset serves exactly once as the test set. Importantly, the classifier remains fixed at the state reached at the end of training in each fold; no further updates or fine-tuning are applied during the testing phase. This evaluation strategy ensures that the statistical properties of the test data closely mirror those of the training data, as both are drawn from the same distribution and processed using identical feature extraction pipelines. As a result, the assessment remains consistent and unbiased across folds.

During testing, each sample’s latent representation---derived from the biological reservoir---is fed into the trained classifier, and the predicted label is compared against the ground truth. Classification performance is quantified using accuracy, defined as the proportion of correctly predicted samples over the total number of test samples. This metric provides a direct and interpretable measure of the system’s ability to generalize across input patterns, offering insight into the effectiveness of the biological reservoir in enabling robust downstream computation.

\section{Experimental Evaluation}
\label{sec:exp}
In this section, we present the experimental setup used to evaluate the proposed BRC system and discuss the results. 
To this end, we designed a progressive series of experiments comprising four levels of stimulus complexity, each encoding distinct spatial input patterns: (i) pointwise stimulations,
(ii) oriented bar patterns, 
(iii) clock-digit-like configurations, and iv) handwritten digits from the MNIST dataset~\cite{726791}. Together, these results allow us to characterize how the representational capacity of the biological reservoir scales with input complexity. For completeness, we also compare the BRC system with an artificial reservoir baseline implemented under matched conditions.

\begin{table}[t]
    \centering
    \footnotesize
    \caption{\textbf{Experimental results for pointwise and oriented-bar stimuli.} 
    Classification accuracy achieved by the BRC system across the two experimental settings: (i) pointwise stimuli and (ii) oriented bars. 
    For each case, we report the mean $\pm$ standard deviation across individual experiments, along with the overall average. 
    For comparison, we also include the performance of an artificial reservoir (AR) evaluated under matched conditions.}
    \vspace{0.5em}
    \renewcommand{\arraystretch}{1.1}
    \setlength{\tabcolsep}{8pt}

    \begin{tabularx}{\linewidth}{
        >{\raggedright\arraybackslash}p{0.34\linewidth} |
        >{\centering\arraybackslash}p{0.24\linewidth} |
        >{\centering\arraybackslash}p{0.24\linewidth} }

        \toprule
        \textbf{Scenario} & \textbf{BR Acc. (\%) $\uparrow$} & \textbf{AR Acc. (\%) $\uparrow$} \\
        \midrule

        \textbf{Pointwise Stimuli} & & \\
        \arrayrulecolor{black!20}\cmidrule(lr){1-3}\arrayrulecolor{black}

        \quad Point 1 & 100\% $\pm$ 0\% & 89\% $\pm$ 13\% \\
        \quad Point 2 & 100\% $\pm$ 0\% & 71\% $\pm$ 19\% \\
        \quad Point 3 & 100\% $\pm$ 0\% & 85\% $\pm$ 18\% \\
        \quad Point 4 & 95\%  $\pm$ 9\% & 79\% $\pm$ 24\% \\
        \quad \textbf{Average} & \textbf{98\% $\pm$ 2\%} & \textbf{82\% $\pm$ 6\%} \\
        \midrule

        \textbf{Oriented Bars} & & \\
        \arrayrulecolor{black!20}\cmidrule(lr){1-3}\arrayrulecolor{black}

        \quad Bar 1 (0 degrees)   & 95\% $\pm$ 9\% & 100\% $\pm$ 0\% \\
        \quad Bar 2 (45 degrees)  & 91\% $\pm$ 11\% & 100\% $\pm$ 0\% \\
        \quad Bar 3 (90 degrees)  & 87\% $\pm$ 27\% & 98\% $\pm$ 3\% \\
        \quad Bar 4 (135 degrees) & 95\% $\pm$ 9\% & 100\% $\pm$ 0\% \\
        \quad \textbf{Average}    & \textbf{92\% $\pm$ 6\%} & \textbf{98\% $\pm$ 3\%} \\
        \bottomrule
    \end{tabularx}

    \label{tab:results_pointwise_bars}
\end{table}

\subsection{Artificial Reservoir Baseline}
To establish a reference point and enable direct comparison, we implemented an artificial reservoir (AR) composed of 4{,}096 recurrently connected rate-based units, matching the dimensionality of the feature vectors extracted from the 4{,}096 electrodes of the HD-MEA. The recurrent connectivity matrix was initialized with 10\% sparsity, consistent with standard artificial-reservoir configurations and inspired by the sparse connectivity observed in cortical circuits. To ensure stable yet expressive dynamics, the spectral radius was set to 0.9, providing a balance between dynamic richness and fading memory. Although our experimental paradigm did not require long-term temporal memory---as reservoir states were sampled immediately after stimulus presentation---this parameter choice yielded consistent and diverse state trajectories.

To mirror our biological experiments as closely as possible, the AR was initialized to a resting state and driven by the same spatially encoded stimulus patterns. We also modeled biologically realistic input variability by superimposing noise derived from empirical measurements: specifically, we sampled $N > 20$ windows of spontaneous activity, computed the mean spike count, and used this distribution to generate the noise injected into the AR input. The resulting reservoir state was then fed into a single-layer perceptron trained to classify the patterns.

It is worth noting that, while the AR provides a useful benchmark, it should be regarded as an approximate upper bound, as it is engineered---by design---to operate under idealized, noise-controlled, and fully optimized conditions and is therefore expected to yield stronger performance. Nevertheless, our BRC system retains the distinctive advantages of a biological substrate, contributing to the broader effort to integrate biological principles into machine learning and offering potential energy-efficiency benefits.

\subsection{Pointwise and Oriented-Bar Experiments}

\subsubsection{Experimental setup}
In this experimental setup, we first applied pointwise stimulation patterns to probe the basic responsiveness of the biological reservoir. These experiments served as a baseline for calibrating the stimulation protocol and validating the feasibility of linear classification. Specifically, four single pairs of adjacent electrodes were activated with opposite polarities (see Fig.~\ref{fig:input_patterns}, first row), selected from regions exhibiting the highest levels of spontaneous spiking activity. Each pair was stimulated using monophasic pulses with amplitude $A = 10\ \mu\mathrm{A}$ and pulse width $\delta = 20\ \mu\mathrm{s}$; the post-stimulus time window $W$ was set to $5\,\si{\milli\second}$.

Then, to increase complexity, we introduced bar-shaped activation patterns to assess the ability of the reservoir to encode spatial structure. Specifically, we defined four stimulus classes, corresponding to bars oriented at 0, 45, 90, and 135 degrees. Each bar consisted of five adjacent electrode pairs aligned along the designated orientation, spaced by one dilation step (see Fig.~\ref{fig:input_patterns}, second row). All bars were centered on the region of highest spontaneous activity, creating overlapping stimulation zones across classes. This design prevents trivial spatial separability, requiring the reservoir to perform meaningful spatiotemporal transformations to produce distinct latent representations. Also in this case, we employed monophasic electrical pulses with the same values of $A$ and $\delta$, as well as $W$.

\begin{figure*}[t]
    \centerline{\includegraphics[width=1\linewidth]{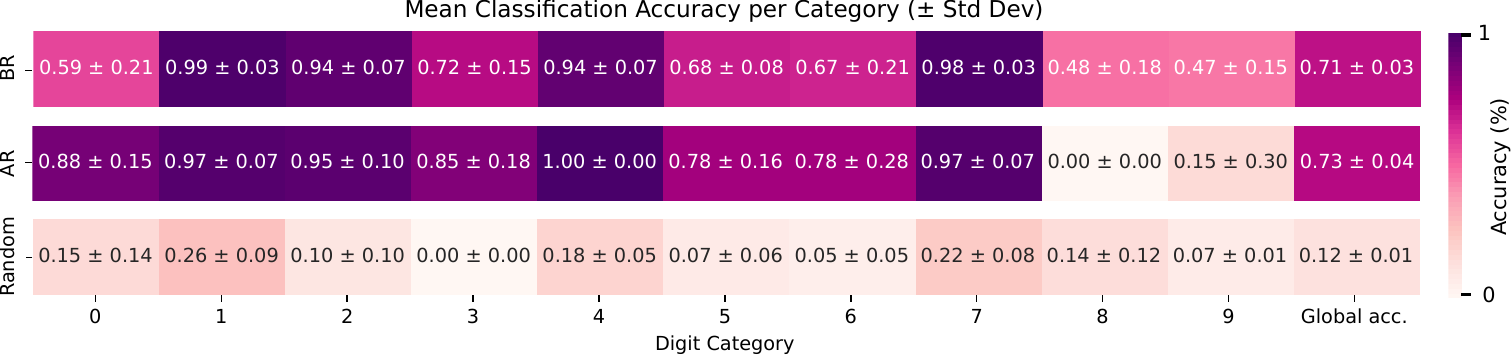}}
    \caption{
    \textbf{Classification performance on clock‑digit‑like patterns.} 
    Mean classification accuracy ($\pm$ standard deviation) for each input category, aggregated over nine independent stimulation sessions. 
    The top row reports results from our biological reservoir computing (BRC) system, the middle row reports the accuracy of the artificial reservoir (AR) averaged across multiple noise realizations, and the bottom row shows the accuracy of a random‑guess baseline obtained by spatially shuffling the neuronal responses from the test days. 
    Overall, the BRC achieves performance comparable to the AR. 
    }
    \label{fig:clk_results}
\end{figure*}

\subsubsection{Quantitative results}
The results are reported in Tab.~\ref{tab:results_pointwise_bars}. 
For the pointwise stimuli, the classifier achieved an accuracy of 98\% $\pm$ 2\%, as expected given the simplicity of the task. The stimuli were spatial patterns that were already linearly separable, which contributed to the high performance obtained by the downstream classifier. Notably, our BRC system outperforms the AR competitor, which struggles despite the simplicity of the setting. We argue that this is due to the way noise is injected in the AR: when applied to pointwise stimuli, the noise perturbs specific input locations, thereby degrading the informative signal. This effect is specific to this experiment, as the subsequent scenarios rely on spatially distributed patterns rather than isolated pointwise activation. While simple, this task confirms that the biological reservoir can respond robustly under controlled, linearly separable conditions.

In the oriented-bars experiment, the introduction of spatially overlapping bar-shaped stimuli significantly increased task complexity. The classifier achieved an average accuracy of 92\% $\pm$ 6\%, representing a modest decrease relative to the pointwise condition. Despite increased overlap in evoked activity, the BRC system continues to produce distinct, linearly separable representations, demonstrating its intrinsic ability to encode and transform spatial patterns even under high interference. As expected, performance was slightly below that of the artificial AR in this more challenging setting, yet the BRC system remained competitive and exhibited meaningful representational structure.

\subsection{Clock-digit-like Experiments}

\subsubsection{Experimental setup}
In this experimental setting, we considered a more complex pattern recognition task: clock-digit-like recognition. In this scenario, patterns shaped like digits were defined using subsets of HD-MEA electrodes. Specifically, we designed ten distinct configurations representing the digits from 0 to 9, each mapped onto the HD-MEA as \textit{electronic clock} patterns. This strategy mimics the layout of a digital clock display, where each subset of electrodes corresponds to an LED segment that becomes active to form the desired digit. These input patterns are shown in the third row of Fig.~\ref{fig:input_patterns}. 
To support this task, we adopted a more complex biphasic stimulation protocol. The stimulus amplitude $A$ was reduced to $\SI{4}{\micro\ampere}$ per electrode pair, which still produced reliable activation thanks to the broader spatial distribution of the stimulation across multiple electrodes. As in the previous scenario, all patterns were delivered within the same region of the HD-MEA grid, chosen for its high level of spontaneous activity and good signal quality.
Regarding the post-stimulus time window $W$, we set it to $5\,\si{\milli\second}$ (see the following ablation study for details). Finally, to obtain statistically robust results, we conducted $n = 9$ independent stimulation sessions across three biological replicates of the same neural culture, each recorded on a different day.

\subsubsection{Quantitative results}
The average classification accuracy for each input category (i.e., for each input digit), computed over the nine stimulation sessions, together with the accuracy of the artificial reservoir (AR) averaged across multiple noise realizations and a random-guess baseline, is summarized in Fig.~\ref{fig:clk_results}\footnote{The results differ slightly from those reported in our previous preliminary work~\cite{10350607} due to the more conservative artifact‑removal protocol adopted in the present study.}. In the random-guess condition, the neuronal responses from the test days were spatially randomized across recording channels: for each stimulus, we preserved the overall response magnitude while shuffling the spatial distribution of spiking activity across electrodes, thereby removing the informative spatial structure while maintaining the activity level.  Our BRC system reaches performance levels comparable to those of the AR, with an average classification accuracy of approximately $71\%$. 
Interestingly, two stimulus categories---specifically, patterns ``1'' and ``7''---consistently achieved higher classification accuracy across all sessions. Although the precise factors underlying this robustness cannot be conclusively identified from the present analysis, one plausible explanation is that these patterns involve fewer stimulation electrodes and show limited spatial overlap with other stimuli. Such characteristics may lead to evoked responses that are more distinct and less affected by interference, thereby facilitating their classification. Nonetheless, this interpretation remains speculative, and additional targeted experiments would be required to rigorously investigate the relationship between stimulus complexity, spatial interactions, and classification stability.

\subsubsection{Ablation studies}
\begin{figure}[t]
    \centerline{\includegraphics[width=0.9\linewidth]{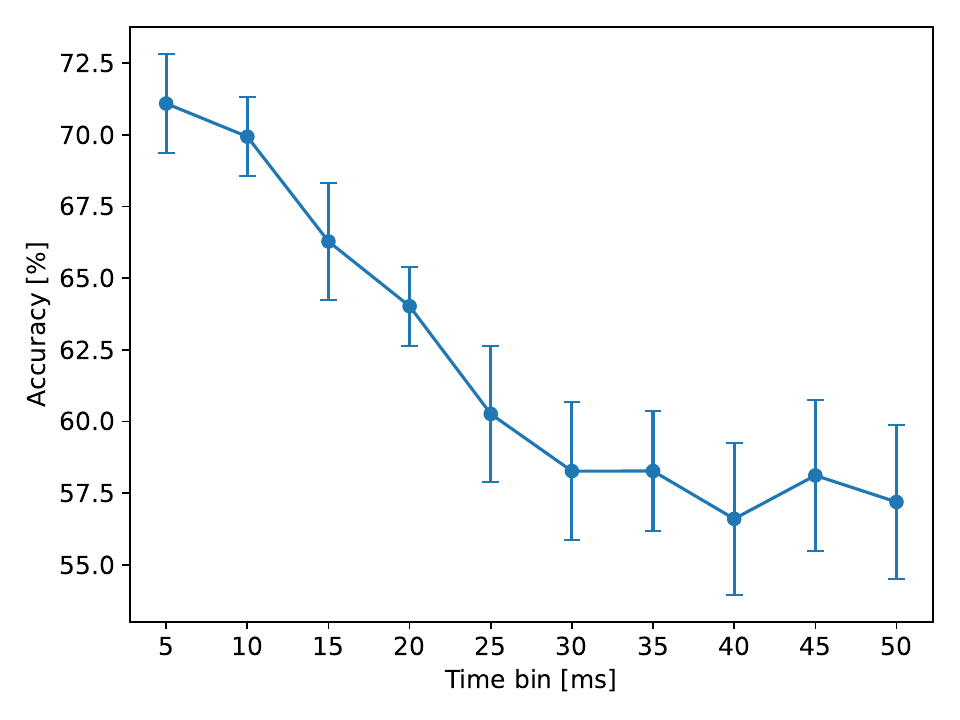}}
    \caption{\textbf{Impact of the readout window duration on classification accuracy in the clock-digit-like experiments.}
    For each of the 9 stimulation sessions, classification accuracy was computed using neural responses elicited by clock-digit-like spatial stimulation patterns and extracted from post-stimulus time windows $W$ ranging from $5$ to $50\,\si{\milli\second}$. 
    The plot reports the mean accuracy together with the standard error of the mean, showing how the duration of the readout window influences the discriminability of the neural representations evoked by clock-digit-shaped inputs.}
    \label{fig:acc_var_w_clk}
\end{figure}

\paragraph{Post-stimulus time window $W$}
To identify the time window that provides the highest amount of information for reading out the response of the network to stimulation, we analyzed the classification accuracy obtained in different stimulation sessions while systematically varying the duration $W$ of the post-stimulus time window used to extract neural activity. This analysis was unnecessary in the previous experiments involving pointwise and oriented‑bar patterns, given the intrinsic simplicity of those stimuli. Figure~\ref{fig:acc_var_w_clk} reports the results aggregated over the $n = 9$ independent stimulation sessions performed across multiple days and biological replicates. A clear decreasing trend emerges, with the highest classification accuracy consistently observed for $W = 5\,\si{\milli\second}$. This indicates that the most discriminative portion of the neural response lies within the first few milliseconds after stimulation.
This finding is consistent with biological considerations. The initial part of the response mainly reflects the immediate activation of neurons directly connected to the stimulated electrodes. In cortical circuits, a window of $5\,\si{\milli\second}$ approximately matches the timescale of a single synaptic transmission and therefore captures the direct postsynaptic potentials triggered by the stimulus. At these short latencies, the signal is minimally affected by recurrent interactions, spontaneous activity, or noise propagation, which makes it highly specific to the input pattern.
When the time window $W$ is increased, later components of the response become progressively influenced by indirect activation, recurrent network dynamics, and ongoing background activity. These contributions reduce the specificity of the recorded signal and explain the drop in classification accuracy observed for larger $W$. These considerations also confirm that the choice of $W = 5\,\si{\milli\second}$ in the previous experiments involving pointwise and bar stimuli was well justified.

\paragraph{Day‑wise performance}
In Tab.~\ref{tab:acc_clk}, we report an ablation of the day‑by‑day performance, presenting both the average accuracy for each stimulation day and the results of the individual stimulation sessions obtained from the three biological replicates. As we can see, the performance shows some variability across days but remains generally stable and comparable overall, demonstrating the robustness of our framework and experimental protocol.

\begin{table}[t]
    \centering
    \footnotesize
    \caption{\textbf{Classification accuracy across stimulation days and biological replicates for the clock‑digit-like experiments.}
    For each stimulation day (Day~1–3), the table reports the classification accuracy obtained from the three biological replicates (BR1–BR3), expressed as mean$\pm$SD over 5‑fold cross‑validation. The final column summarises the day‑wise performance by averaging across replicates.}
    \vspace{0.5em}
    \renewcommand{\arraystretch}{1.15}
    \setlength{\tabcolsep}{3pt}

    \begin{tabularx}{\linewidth}{@{}%
        l
        !{\vrule width 0.5pt}
        >{\centering\arraybackslash}p{0.16\linewidth}
        >{\centering\arraybackslash}p{0.16\linewidth}
        >{\centering\arraybackslash}p{0.16\linewidth}
        !{\vrule width 0.5pt}
        >{\centering\arraybackslash}X
        @{}}
        \toprule
        \textbf{Stim. Day} & \multicolumn{3}{c!{\vrule width 0.5pt}}{\textbf{Acc. per Bio. Rep. (\%) $\uparrow$}} & \textbf{Avg. Day Acc. (\%) $\uparrow$} \\
        \cmidrule(l){2-4}
         & \textbf{BR1} & \textbf{BR2} & \textbf{BR3} & \\
        \midrule
        Day 1 & 74\% $\pm$ 6\% & 70\% $\pm$ 9\% & 71\% $\pm$ 4\% & 72\% $\pm$ 7\% \\
        Day 2 & 72\% $\pm$ 9\% & 75\% $\pm$ 8\% & 71\% $\pm$ 9\% & 73\% $\pm$ 8\% \\
        Day 3 & 70\% $\pm$ 4\% & 73\% $\pm$ 9\% & 64\% $\pm$ 5\% & 69\% $\pm$ 7\% \\
        \bottomrule
    \end{tabularx}

    \label{tab:acc_clk}
\end{table}

\begin{figure*}[t]
    \centerline{\includegraphics[width=1\linewidth]{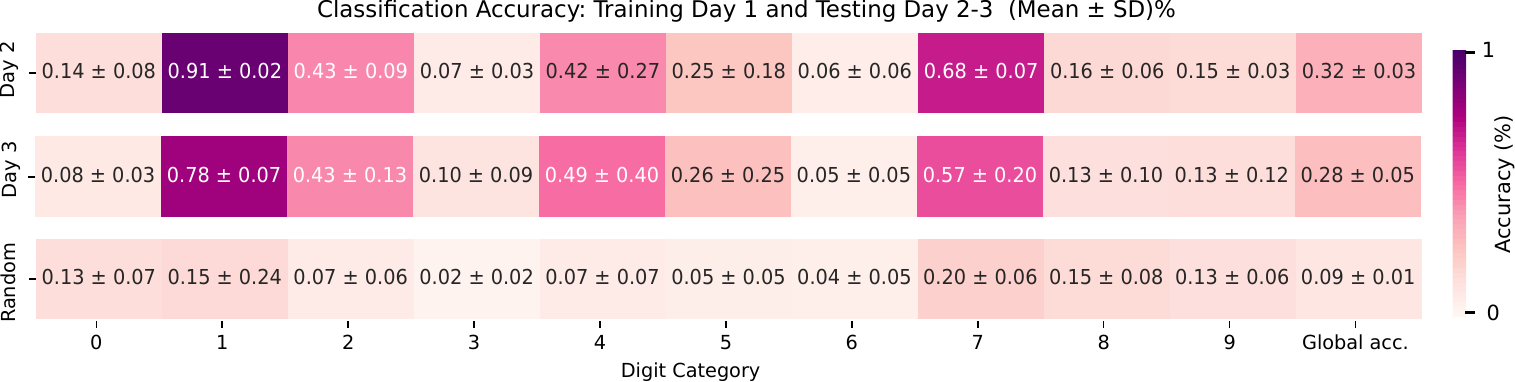}}
    \caption{
    \textbf{Cross‑day generalization performance on clock‑digit‑like patterns.}
    Mean classification accuracy ($\pm$ standard deviation) obtained when training the readout on Day~1 and testing on stimulation sessions recorded on Day~2 and Day~3. 
    Each row reports the accuracy for one test day across all digit categories, while the bottom row shows the performance of a random‑guess baseline. 
    Classification accuracy progressively declines across days, indicating substantial reorganization of the underlying neural dynamics over time; however, all values remain above the simulated chance level, suggesting partial preservation of stimulus‑evoked structure despite ongoing developmental drift in the biological reservoir.
    }
    \label{fig:clk_results_pretrain}
\end{figure*}

\paragraph{Cross-day performance}
Figure~\ref{fig:clk_results_pretrain} reports the results obtained in the cross-day setting, compared with the random guess baseline. In this configuration, the classifier was trained using the three stimulation sessions recorded from the biological replicates of the same culture on the first day (Day~1), and subsequently tested on the six stimulation sessions collected from the replicates measured on the following days (Day~2 and Day~3). This cross-day evaluation probes the extent to which the neural dynamics elicited by a given set of stimuli remain stable over time, or whether they undergo spontaneous drift and reorganization, thus affecting the stability and generalizability of stimulus-evoked responses.
Indeed, neural cultures are known to evolve dynamically across days, transitioning through distinct activity regimes---from uncoordinated, irregular spiking to increasingly synchronized bursting patterns~\cite{fabri, IANNELLO2025116184}. Such reorganization, driven by intrinsic developmental processes and continual remodeling of synaptic connectivity, can substantially alter how the network responds to identical stimulation patterns. As a result, stimulus-evoked responses may vary markedly from one day to the next, reflecting shifts in the functional state of the culture.
The results show a progressive degradation in classification performance across days, with accuracy decreasing to $46 \pm 15\%$ on Day~2 and further to $37 \pm 19\%$ on Day~3. Even though both values remain above the simulated chance level (12\%), this decline indicates that the underlying neural dynamics and connectivity patterns within the biological reservoir undergo substantial reconfiguration across days, likely driven by spontaneous activity and intrinsic plasticity mechanisms typical of neuronal cultures. These observations provide an important insight for future work, suggesting that strategies aimed at stabilizing or compensating for such temporal variability could enhance the long-term robustness of biological reservoir computing.

\begin{figure}[t]
    \centerline{\includegraphics[width=1\linewidth]{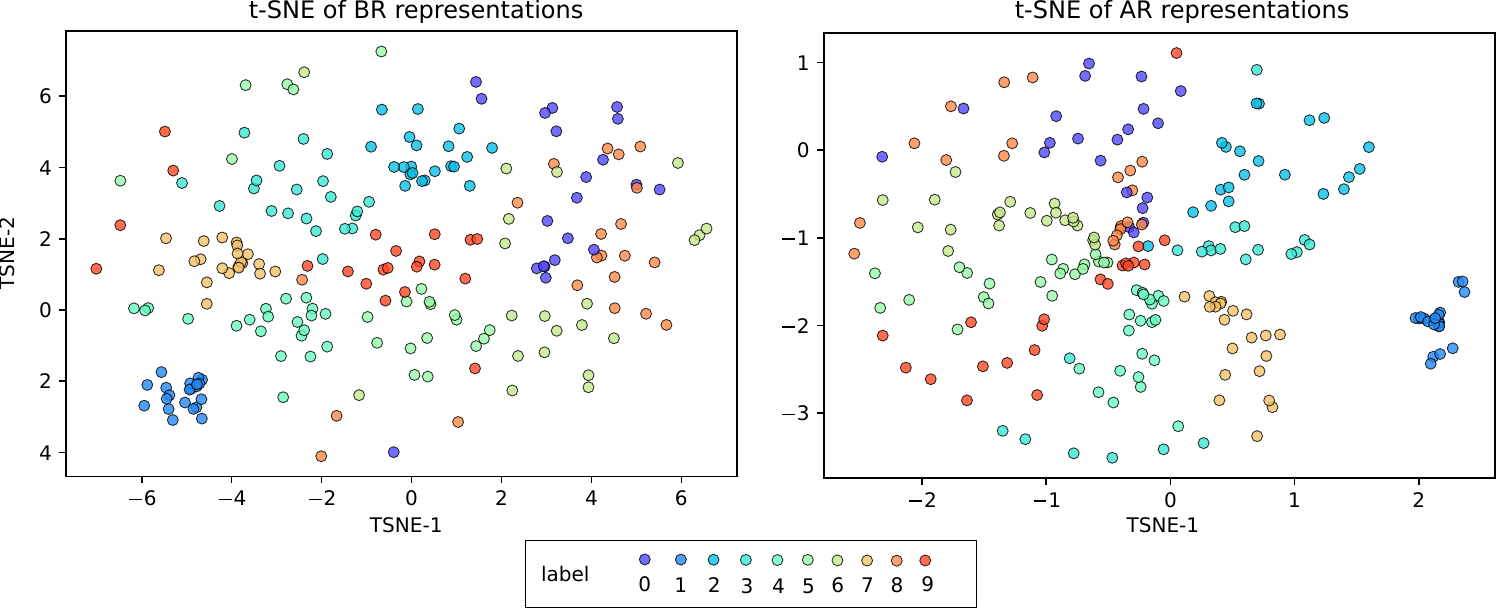}}
    \caption{\textbf{t-SNE low-dimensional projections of reservoir state representations concerning the clock-digit-like experiments}.  
    Left: representations generated by our biological reservoir.  
    Right: representations generated by the artificial reservoir.  
    Each point corresponds to the reservoir response to a single-digit stimulus, with colors indicating digit classes. The BR produces broader and more overlapping structures, reflecting its intrinsic variability and the distributed nature of its encoding. In contrast, the AR yields more compact and clearly separated clusters. These visualizations provide a qualitative comparison of the geometric organization of the representations generated by biological and artificial reservoirs.}
    \label{fig:clock_low_dimensional_projections}
\end{figure}


\paragraph{Low-dimensional representation of BR states}
Figure~\ref{fig:clock_low_dimensional_projections} summarizes the geometric organization of the reservoir state representations obtained using our biological reservoir. Specifically, the figure displays the t-SNE low-dimensional projections~\cite{NIPS2002_6150ccc6} of the high-dimensional reservoir states elicited by the clock-digit-like inputs. Each point corresponds to the reservoir response to a single stimulation instance, and colors denote the digit classes.
Even in the absence of any supervised learning within the reservoir, the induced neural dynamics give rise to structured, distributed representations in which different digit classes tend to occupy partially separable regions of the state space. At the same time, the projections reveal substantial overlap between classes, which is expected given the intrinsic variability, stochasticity, and recurrent interactions characteristic of biological reservoirs. Rather than forming clearly segregated clusters, the responses are arranged into partially overlapping structures whose discriminability ultimately emerges at the level of the linear readout.
A qualitative comparison with the AR (right panel) highlights a marked difference in the geometry of the induced representations: while the artificial reservoir produces more compact, clearly separated clusters (as expected), the BR yields broader, more overlapping structures, consistent with its higher intrinsic variability. Importantly, the presence of such overlap indicates a partial reduction of discriminative information within the biological reservoir. However, this reduction appears limited, as evidenced by the classification performance reported in Fig.~\ref{fig:clk_results}. This suggests that the biological reservoir still encodes stimulus-specific structure in its activity, although in a more diffuse and distributed form.

\subsection{Handwritten Digit MNIST Experiments}

\subsubsection{Experimental setup}
\begin{figure}[t]
    \centerline{\includegraphics[width=0.95\linewidth]{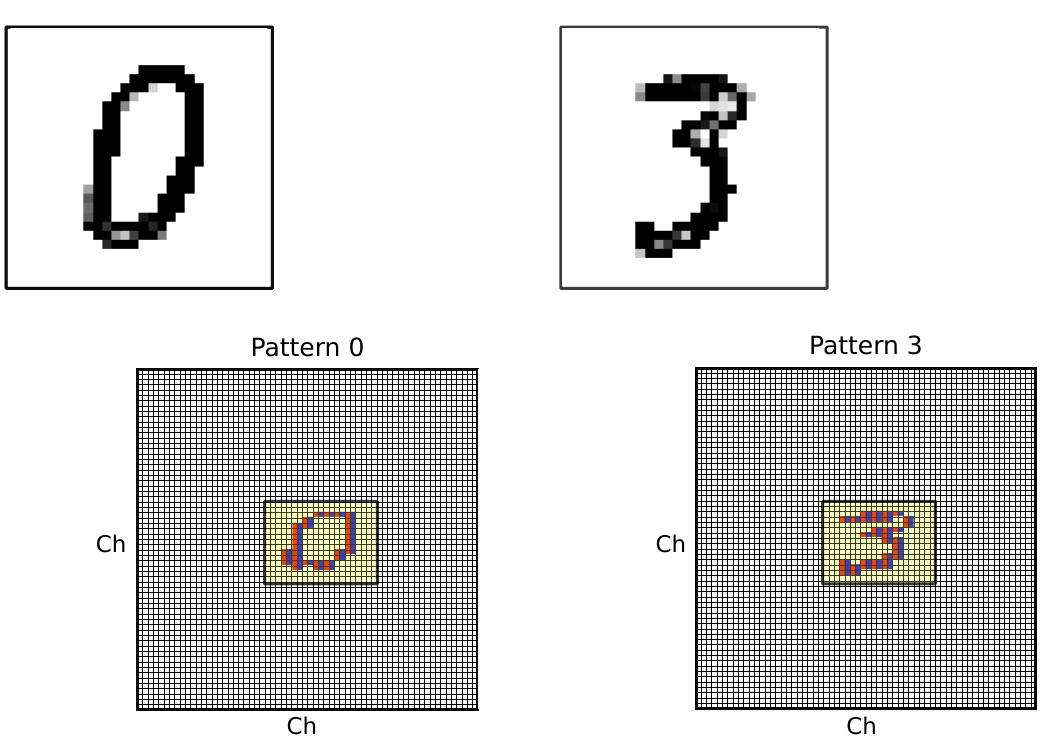}}
    \caption{\textbf{Mapping from MNIST images to HD‑MEA stimulation electrodes.} The figure depicts the transformation of handwritten digit images into stimulation patterns for the HD‑MEA, illustrated with two samples corresponding to the digits 0 and 3 and used for handwritten digit classification.}
    \label{fig:mapping_mnist}
\end{figure}

In this setup, we considered the widely used MNIST dataset~\cite{726791}, moving toward the more challenging task of handwritten digit recognition. In our experiments, we employed a randomly sampled subset of 200 images from the dataset, which was sufficient to probe the ability of the biological reservoir to generalize across digit categories while keeping the stimulation protocol manageable. Specifically, each MNIST image (28$\times$28 pixels, grayscale) is first downsampled to a lower resolution using nearest‑neighbor interpolation to match the spatial layout of the HD-MEA and to avoid issues during stimulation. The resulting image is then stretched horizontally to ensure that each pixel is associated with a pair of adjacent electrodes, one positive and one negative, which are required for stimulation. No thresholding is applied to binarize the images. Instead, the grayscale intensity of each pixel directly modulates the probability of stimulating the corresponding electrode. Darker pixels (i.e., lower intensity values) are associated with a higher probability of triggering a stimulation event. When stimulation occurs at a given pixel location, a pair of electrodes is activated: the electrode corresponding to the pixel position acts as the positive pole, while an adjacent electrode (e.g., to the right) serves as the negative pole. This design produces spatially localized, contrast‑dependent stimulation patterns that preserve the structural information of the original input. An illustration of the mapping procedure from two MNIST samples representing the digits 0 and 3 to the HD‑MEA electrode layout is shown in Fig.~\ref{fig:mapping_mnist}; in addition, ten representative input patterns, each corresponding to a different digit, are reported in Fig.~\ref{fig:input_patterns} (row~4). As in the previous set of experiments on clock‑digit‑like classification, we adopted biphasic stimulation with an amplitude of $A = \SI{5}{\micro\ampere}$. Consistent with earlier experiments, all digit patterns were delivered within the same region of the HD‑MEA grid, selected for its high level of spontaneous activity and good signal quality. Regarding the post‑stimulus time window $W$, we set it to $10\,\si{\milli\second}$ (see the following ablation study for details). Finally, as in the previous set of experiments, to ensure statistically robust results, we conducted $n = 9$ independent stimulation sessions across three biological replicates of the same neural culture, each recorded on a different day.

\begin{figure*}[t]
    \centerline{\includegraphics[width=1\linewidth]{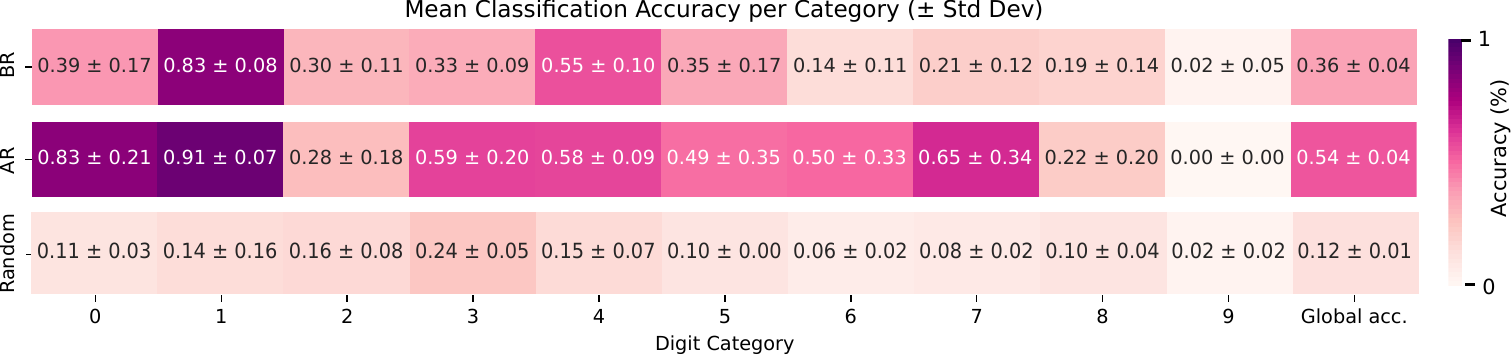}}
    \caption{
    \textbf{Classification performance on MNIST handwritten digits.}
    Mean classification accuracy ($pm$ standard deviation) for each digit category, aggregated over nine stimulation sessions, for our BRC system, the AR, and the random‑guess baseline.  
    The BRC achieves accuracy well above the random baseline and, for several digits, reaches performance levels comparable to the AR, demonstrating its ability to classify complex handwritten patterns. The rightmost column reports the global accuracy averaged across all digit categories.
    }
    \label{fig:mnist_results}
\end{figure*}

\subsubsection{Quantitative results}
Figure~\ref{fig:mnist_results} reports the average classification accuracy obtained for each input digit across the nine stimulation sessions, and compares it with both the average AR accuracy measured under different noise realizations and a random‑guess baseline. Our BRC system achieves performance far above the random‑guess baseline and, for several digits, reaches accuracy levels comparable to the AR model, demonstrating once again its ability to successfully classify complex spatial patterns. It is worth noting that digit ``9'' performs worse in all scenarios; we attribute this to the fact that the dataset is not perfectly balanced, and the pattern corresponding to digit ``9'' is under‑represented in the training samples.

\subsubsection{Ablation studies}

\paragraph{Post-stimulus time window $W$}
As in the previous set of experiments, we performed an ablation study in which we varied the duration of the post‑stimulus time window $W$ used to extract neural activity. A clear decreasing trend emerges after $W = 10\,\mathrm{ms}$, where the highest classification accuracy is observed. We deem that the reason why the optimal window increased from $5\,\mathrm{ms}$ in the previous experiments to $10\,\mathrm{ms}$ here is that in the present study we stimulated a substantially larger number of electrodes. As a consequence, the evoked responses were stronger and propagated more broadly across the network. Under these conditions, a $10\,\mathrm{ms}$ window---approximately matching the timescale of one to two synaptic transmissions---captures more of the informative, stimulus‑driven activity, thereby improving classification accuracy.

\begin{figure}[t]
    \centerline{\includegraphics[width=0.9\linewidth]{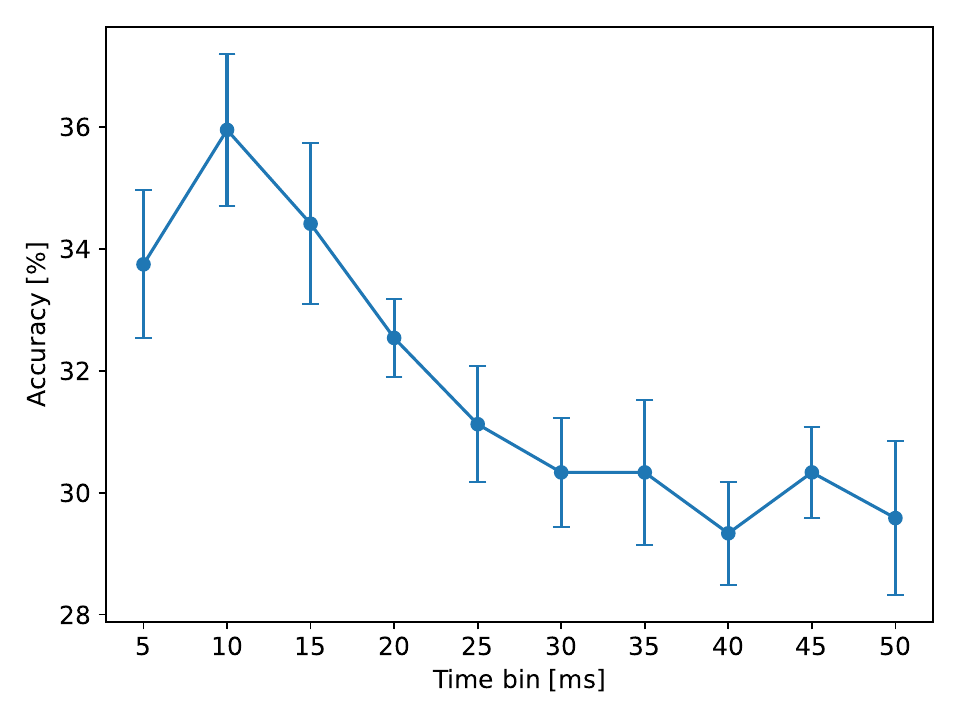}}
    \caption{\textbf{Impact of the readout window duration on classification accuracy in the MNIST handwritten digit experiments.} 
    For each of the nine stimulation sessions, classification accuracy was computed from neural responses elicited by the MNIST digit‑like spatial stimulation patterns and extracted using post‑stimulus time windows \(W\) ranging from \(5\) to \(50\,\si{\milli\second}\). 
    The plot shows the mean accuracy together with the standard error of the mean, illustrating how the duration of the readout window shapes the discriminability of the evoked neural representations.}
    \label{fig:acc_var_w}
\end{figure}

\paragraph{Day‑wise performance}
As in the previous set of experiments, Tab.~\ref{tab:acc_mnist} reports an ablation of the day-wise performance, presenting both the average accuracy for each stimulation day and the results of the individual stimulation sessions obtained from the three biological replicates. In this case as well, the performance shows some variability across days but remains generally stable and comparable overall, demonstrating the robustness of our framework and experimental protocol.

\begin{table}[t]
    \centering
    \footnotesize
    \caption{\textbf{Classification accuracy across stimulation days and biological replicates for the MNIST handwritten digit experiments.}
    For each stimulation day (Day~1–3), the table reports the classification accuracy obtained from the three biological replicates (BR1–BR3), expressed as mean$\pm$SD over 5‑fold cross‑validation. The final column summarises the day‑wise performance by averaging across replicates.}
    \vspace{0.5em}
    \renewcommand{\arraystretch}{1.15}
    \setlength{\tabcolsep}{3pt}

    \begin{tabularx}{\linewidth}{@{}%
        l
        !{\vrule width 0.5pt}
        >{\centering\arraybackslash}p{0.16\linewidth}
        >{\centering\arraybackslash}p{0.16\linewidth}
        >{\centering\arraybackslash}p{0.16\linewidth}
        !{\vrule width 0.5pt}
        >{\centering\arraybackslash}X
        @{}}
        \toprule
        \textbf{Stim. Day} & \multicolumn{3}{c!{\vrule width 0.5pt}}{\textbf{Acc. per Bio. Rep. (\%) $\uparrow$}} & \textbf{Avg. Day Acc. (\%) $\uparrow$} \\
        \cmidrule(l){2-4}
         & \textbf{BR1} & \textbf{BR2} & \textbf{BR3} & \\
        \midrule
        Day 1 & 34\% $\pm$ 9\% & 37\% $\pm$ 5\% & 41\% $\pm$ 7\% & 37\% $\pm$ 8\% \\
        Day 2 & 34\% $\pm$ 9\% & 36\% $\pm$ 8\% & 31\% $\pm$ 5\% & 34\% $\pm$ 8\% \\
        Day 3 & 33\% $\pm$ 6\% & 36\% $\pm$ 8\% & 30\% $\pm$ 6\% & 33\% $\pm$ 7\% \\
        \bottomrule
    \end{tabularx}

    \label{tab:acc_mnist}
\end{table}

\begin{figure*}[t]
    \centerline{\includegraphics[width=1\linewidth]{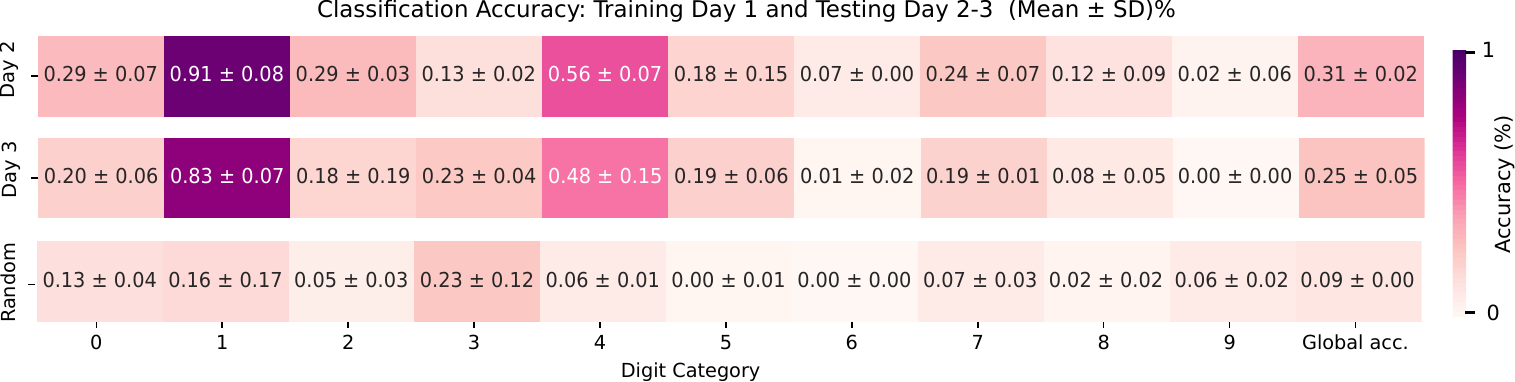}}
    \caption{
    \textbf{Cross‑day generalization performance on MNIST handwritten digit experiments.}
    Mean classification accuracy ($\pm$ standard deviation) when training the readout on Day~1 and testing on stimulation sessions recorded on Day~2 and Day~3. 
    Each row reports the accuracy for one test day across all digit categories, while the bottom row shows a random‑guess baseline. 
    Classification accuracy declines moderately across days, in line with the ongoing maturation and variability of the cultured network.  
    Nevertheless, both values remain well above the simulated chance level, indicating that the culture continues to encode stimulus‑related information despite day‑to‑day shifts in its functional state.
    }
    \label{fig:mnist_results_pretrain}
\end{figure*}

\paragraph{Cross-day performance}
Figure~\ref{fig:mnist_results_pretrain} reports the results obtained in the cross-day setting as in the previous set of experiments. The results show, in general, a progressive degradation in classification performance across days, as expected, with a global accuracy decreasing to $32 \pm 3\%$ on Day~2 and further to $28 \pm 5\%$ on Day~3. However, in this set of experiments as well, both values remain above the simulated chance level ($9 \pm 1\%$) by a large margin.

\begin{figure}[t]
    \centerline{\includegraphics[width=1\linewidth]{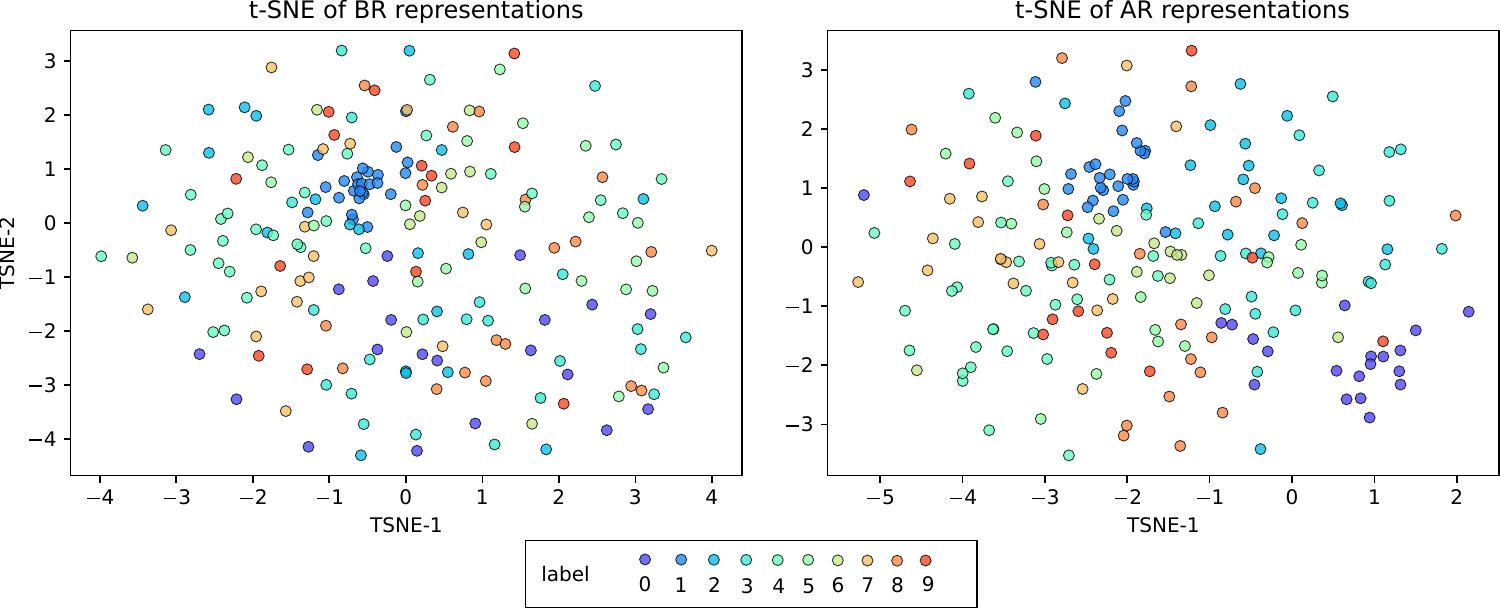}}
    \caption{\textbf{t-SNE low-dimensional projections of reservoir state representations concerning the MNIST handwritten digit experiments}.  
    The left panel shows the representations produced by the biological reservoir, whereas the right panel displays those generated by the artificial reservoir. 
    Each point corresponds to the reservoir state evoked by a single-digit stimulus, with colors indicating the associated digit class. 
    The BR yields more diffuse and partially overlapping clusters, consistent with its higher intrinsic variability and the distributed nature of its encoding. 
    Conversely, the AR produces tighter and more clearly separated clusters. 
    Taken together, these projections provide a qualitative view of how biological and artificial reservoirs differ in the geometric organization of their internal representations.}
    \label{fig:mnist_low_dimensional_projections}
\end{figure}

\paragraph{Low-dimensional representation of BR states}
As in the previous set of experiments, we characterized the geometric organization of the reservoir state representations produced by our biological reservoir by visualizing the low‑dimensional structure of the neural responses. To this end, we applied t‑SNE~\cite{NIPS2002_6150ccc6} to the high‑dimensional reservoir states elicited by the digit inputs (see Fig.~\ref{fig:mnist_low_dimensional_projections}). Each point in the projection corresponds to the reservoir response to a single stimulation instance, and colors denote the digit classes.
Also in this case, the induced neural dynamics give rise to distributed and partially separable representations: different digit classes tend to occupy distinct regions of the state space, although with broader and more overlapping structures compared to artificial reservoirs. This increased spread is consistent with the higher intrinsic variability of the biological substrate (left panel), while artificial reservoirs (right panel) produce more compact and sharply clustered embeddings. Nevertheless, the reduction in cluster compactness observed in the biological case remains limited, as confirmed by the classification performance.

\section{Conclusion and Future Directions}
\label{sec:conclusions}
In this work, we introduced a biological reservoir computing (BRC) framework in which a network of cultured biological neurons serves as the computational substrate. We investigated its applicability to static pattern-recognition tasks, where spatially encoded stimuli elicit transient neural responses that are subsequently used for classification. In contrast to conventional artificial reservoirs, our bio-hybrid approach leverages the intrinsic dynamics of living neuronal circuits, which naturally exhibit energy-efficient operation, rich nonlinear behavior, and adaptive responses. These properties highlight biological reservoirs as a promising direction for neuromorphic computation, particularly in scenarios requiring low-power processing and inherent adaptability. At the same time, this line of research offers a unique opportunity to probe how biological neural networks transform and encode information, potentially contributing to a deeper understanding of neural computation and cognition.

To implement our BRC system, we employed a high-density multi-electrode array (HD-MEA), enabling both electrical stimulation and high-resolution recording of neural activity. The biological substrate consisted of cortical neurons derived \textit{in vitro} from mouse embryonic stem cells, which self-organize into spontaneously active neuronal networks upon maturation. Input patterns were encoded as spatially distributed stimulation sequences, designed to mimic simplified visual stimuli, and delivered to the culture through the HD-MEA. The resulting neural responses were recorded and converted into feature vectors for downstream classification, which was carried out by a simple readout implemented as a single-layer perceptron.

Having established the experimental platform, we systematically evaluated the ability of the system to handle inputs of increasing complexity through a sequence of tasks progressing from simple to challenging patterns. We began with elementary pointwise inputs, used to probe the basic responsiveness of the reservoir. We then introduced structured geometric patterns, such as oriented bars. Next, we assessed the classification of ten clock‑digit‑like stimulation patterns (0–9). Finally, we evaluated the system on a substantially more demanding task: the classification of handwritten digits from the MNIST dataset~\cite{726791}, where each image was mapped to a corresponding spatial stimulation sequence. Despite the intrinsic variability of biological neuronal responses---arising from noise, spontaneous activity, and differences across sessions---the system consistently produced high‑dimensional representations that supported accurate classification. These results demonstrate that cortical neuronal networks cultured \textit{in vitro} can function effectively as reservoirs for static pattern recognition, highlighting their potential as living computational substrates within neuromorphic architectures.

Looking ahead, we plan to extend this work by addressing more complex pattern-recognition tasks, enabling a broader assessment of the scalability and robustness of the biological reservoir across different computational settings. Another promising direction involves integrating optogenetic stimulation protocols~\cite{meloni2020} alongside traditional electrical stimulation. Such hybrid approaches may help mitigate electrode degradation while improving the precision and reliability of network activation, ultimately contributing to more sustainable and flexible neural interfacing strategies. 
Furthermore, exploring stimulation paradigms capable of inducing synaptic plasticity will allow us to investigate how intrinsic learning mechanisms can be harnessed to shape the functional properties of the biological reservoir, potentially guiding the network toward more efficient or task-specialized responses. Together, these future research avenues hold significant promise for advancing both the fundamental understanding and the practical deployment of biological reservoir computing, moving us closer to energy-efficient and biologically grounded computational systems.

\section*{Acknowledgements and Funding}
This work has been supported by the Matteo Caleo Foundation, by Scuola Normale Superiore (FC), by the PRIN project ''AICult'' (grant \#2022M95RC7) from the Italian Ministry of University and Research (MUR) (FC), and by the PNRR project ``Tuscany Health Ecosystem - THE'' (CUP B83C22003930001) funded by the European Union - NextGenerationEU. 

\section*{Data Availability}
The MNIST dataset is publicly available at~\cite{726791}.


\bibliographystyle{elsarticle-num} 
\bibliography{biblio.bib}



\end{document}